\documentclass[conference]{IEEEtran}

\ifCLASSINFOpdf
\else
\fi
\usepackage{amsmath,graphicx}

\usepackage{graphicx}
\usepackage{amsmath}
\usepackage{amssymb}
\usepackage{booktabs}
\usepackage{url}

\usepackage{comment}
\usepackage{multirow}
\usepackage{times}
\usepackage{epsfig}
\usepackage{graphicx}
\usepackage{amsmath}
\usepackage{amssymb}
\usepackage{booktabs}
\usepackage{xcolor}
\usepackage{color}

\begin{document}

\title{TransNet: A Transfer Learning-Based Network for \\ Human Action Recognition}

\author{\IEEEauthorblockN{Khaled Alomar}
\IEEEauthorblockA{School of Electronics and Computer Science\\ University of Southampton, SO17 1BJ, Southampton, UK\\
Email:  k.a.alomar@soton.ac.uk}
\and
\IEEEauthorblockN{Xiaohao Cai}
\IEEEauthorblockA{School of Electronics and Computer Science\\ University of Southampton, SO17 1BJ, Southampton, UK\\
Email:  x.cai@soton.ac.uk}}


%Telephone: (800) 555--1212, Fax: (888) 555--1212}


\maketitle

\begin{abstract}
Human action recognition (HAR) is a high-level and  significant research area in computer vision due to its ubiquitous applications. The main limitations of the current HAR models are their complex structures and lengthy training time. In this paper, we propose a simple yet versatile and effective end-to-end deep learning architecture, coined as {\it TransNet}, for HAR. 
TransNet decomposes the complex 3D-CNNs into 2D- and 1D-CNNs, where the 2D- and 1D-CNN components extract spatial features and temporal patterns in videos, respectively. Benefiting from its concise architecture, TransNet is ideally compatible with any pretrained state-of-the-art 2D-CNN models in  other fields, being transferred to serve the HAR task.
In other words, it naturally leverages the power and success of transfer learning for HAR, bringing huge advantages in terms of efficiency and effectiveness.
Extensive experimental results and the comparison with the state-of-the-art models demonstrate the superior performance of the proposed TransNet in HAR in terms of flexibility, model complexity, training speed and classification accuracy.
\end{abstract}


% \fi
\IEEEpeerreviewmaketitle
\section{Introduction}
\label{sec:intro}

The computer vision community has studied video analysis for decades, including action recognition \cite{tran2015learning} and activity understanding \cite{kitani2012activity}. Human action recognition (HAR) analyses and detects actions from unknown video sequences. Due to the rising demand for automated behaviour interpretation, HAR has gained dramatic attention from academics and industry and is crucial for many applications \cite{paul2014survey}.

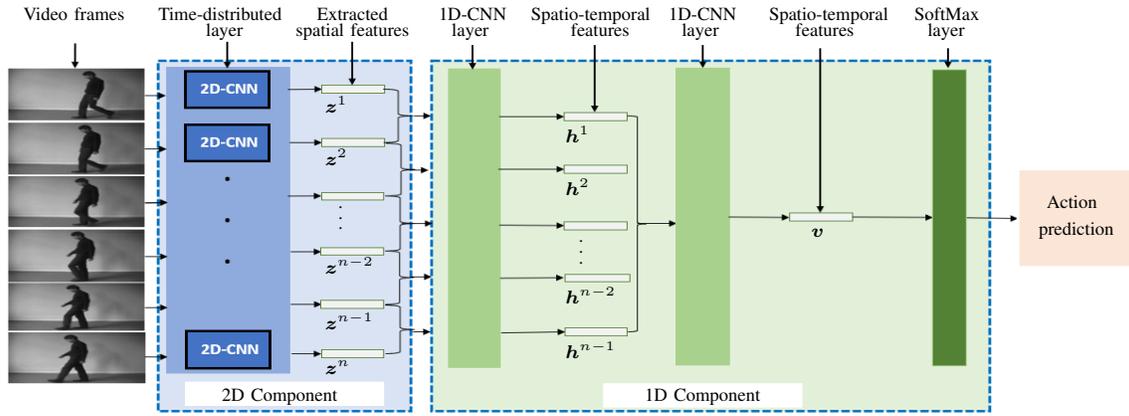
\begin{figure*}[!htb]
    \centering
    \includegraphics[height=2.2in,width=6in]{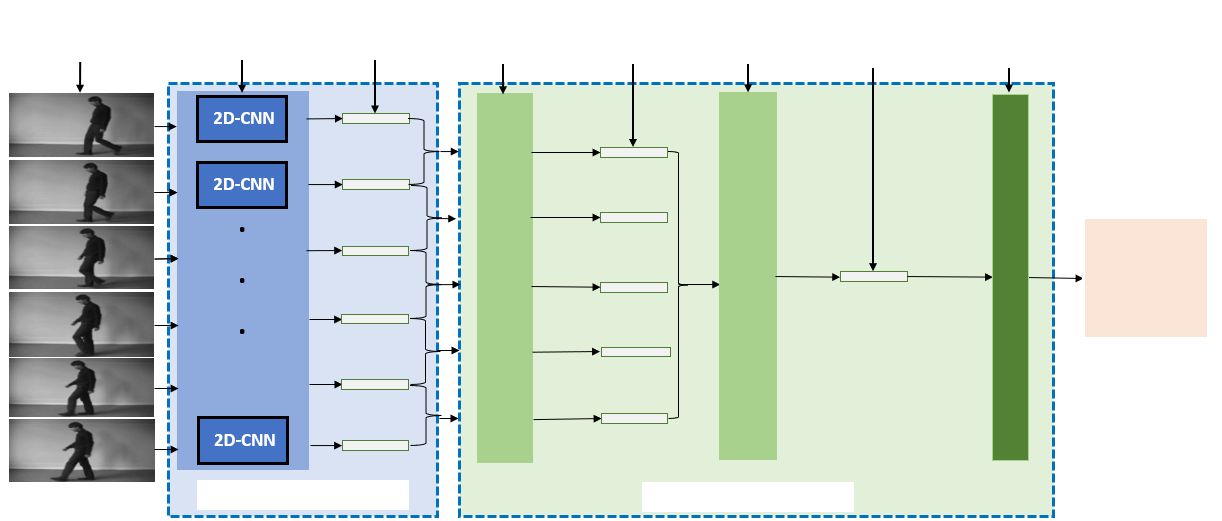}
    \put(-425,149){\scriptsize Video frames}
    \put(-374,149){\scriptsize Time-distributed }
    \put(-356,143){\scriptsize  layer}
    \put(-314,149){\scriptsize Extracted }
    \put(-322,143){\scriptsize spatial features}
    \put(-268,149){\scriptsize 1D-CNN   }
    \put(-263,143){\scriptsize layer}
    \put(-233,149){\scriptsize Spatio-temporal  }
    \put(-218,143){\scriptsize features}
    \put(-181,149){\scriptsize 1D-CNN }
    \put(-176,143){\scriptsize layer}
     \put(-143,149){\scriptsize Spatio-temporal }
    \put(-133,143){\scriptsize features}
    \put(-88,149){\scriptsize SoftMax }
     \put(-83,143){\scriptsize  layer}
    \put(-38,78){\scriptsize Action }
    \put(-41,68){\scriptsize  prediction}
    \put(-190,5){\scriptsize 1D Component }
    \put(-350,5){\scriptsize 2D Component }
    \put(-311,114){\scriptsize $\boldsymbol{z}^{1}$}
    \put(-311,94){\scriptsize $\boldsymbol{z}^{2}$}
    \put(-306,70){\scriptsize $\vdots$}
    \put(-311,53){\scriptsize $\boldsymbol{z}^{n-2}$}
    \put(-311,32){\scriptsize $\boldsymbol{z}^{n-1}$}
    \put(-311,15){\scriptsize $\boldsymbol{z}^n$}
    \put(-220,103){\scriptsize $\boldsymbol{h}^1$}
    \put(-220,83){\scriptsize $\boldsymbol{h}^{2}$}
    \put(-215,57){\scriptsize $\vdots$}
    \put(-220,42){\scriptsize $\boldsymbol{h}^{n-2}$}
    \put(-220,21){\scriptsize $\boldsymbol{h}^{n-1}$}
    \put(-127,67){\scriptsize $\boldsymbol{v}$}
    \caption{TransNet architecture for HAR. The given video frames are input into the time-distributed layer, which employs a 2D-CNN model (e.g., MobileNet, MobileNetV2, VGG16, or VGG19) several times based on the number of video frames, allowing the architecture to analyse multiple frames without expanding in size. Then the spatial features corresponding to the individual input frames are generated, which are subsequently analysed by the 1D-CNN layers, extracting the spatio-temporal features. The SoftMax layer finally classifies the action according to the spatio-temporal pattern.}
    \label{fig:2D+1DModel}
\end{figure*}

Good action recognition requires extracting spatial features from the sequenced frames (images) of a video and then establishing the temporal correlation (i.e., temporal features) between these spatial features. Thus, action recognition models analyse two types of features, establish their relationship, and classify complex patterns. This makes these models vulnerable to a number of significant challenges, including i) the limited ability to transfer learning exploiting advanced models from other fields in computer vision, ii) the need for large volumes of data due to the model complexity, iii) the need for accurate temporal analysis of spatial features, and iv) the overlap of moving object data with cluttered background data \cite{jegham2020vision}.

The improvement process across generations of these models is inconsistent \cite{simonyan2014two}. This results in a wide range of works that may face difficulty of transferring learning ability between generations, especially when these models are constructed differently and/or developed in different fields for extracting specific spatial features in HAR.

\begin{figure}
    \centering
 \includegraphics[scale=0.33]{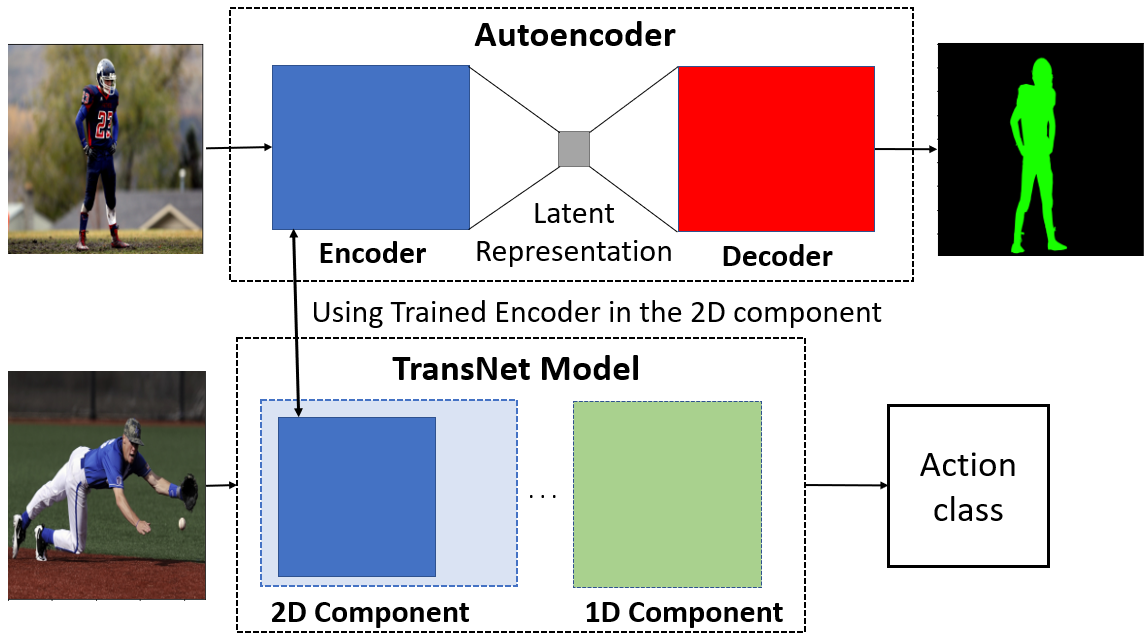}
    \caption{An illustration of TransNet+ for HAR. TransNet+ inherits the architecture of TransNet. It uses the autoencoder's encoder to form the TransNet's 2D component. }
  \label{fig:BypassingProcess}
\end{figure}

Temporal modeling presents a big challenge in action recognition. To address this, researchers often employ 3D-CNN models, which excel at interpreting spatio-temporal characteristics but suffer from much larger size compared to 2D-CNN models \cite{yang2019asymmetric}. Moreover, optimising 3D networks becomes difficult when dealing with insufficient data \cite{kong2021spatiotemporal}, since training a 3D convolutional filter necessitates a substantial dataset encompassing diverse video content and action categories \cite{hu20213d}. Unlike recurrent neural networks (RNNs) that emphasise temporal patterns \cite{narang2017exploring}, 3D networks analyse videos as 3D images, potentially compromising the sequential analysis of temporal data. Both 3D-CNNs and RNNs are challenged by the increased model size and lengthy training time \cite{stamoulakatos2021comparison}.

The presence of cluttered backgrounds presents another challenge in HAR. Indoor environments with static and constant backgrounds are typically assumed to yield high performance for HAR models, whereas performance could significantly diminish in outdoor contexts \cite{liu2015single, wu2011action}. Cluttered backgrounds introduce interruptions and background noise, encoding problematic information in the extraction of global features and leading to a notable decline in performance. To address this challenge, a practical approach is to design models that focus on the human object rather than the background. Scholarly literature consistently indicates that incorporating multiple input modalities, including optical flow and body part segmentation, shows promise in enhancing HAR performance. This conclusion is substantiated by a range of survey studies conducted in the field of action recognition, providing robust evidence for the effectiveness of leveraging diverse input modalities \cite{beddiar2020vision, kong2022human, sun2022human}.

However, there are several issues with these types of models, including their various modelling steps, preprocessing stages, lengthy training time, and significant demands on resources such as memory and processing power. These models are also difficult to implement in real-world applications.

In this paper, we propose an end-to-end deep learning architecture called {\it TransNet} for HAR, see Figure \ref{fig:2D+1DModel}. Rather than using complex 3D-CNNs, TransNet consists of 2D- and 1D-CNNs that extract spatial features and temporal patterns in videos, respectively. TransNet offers the following multiple benefits: i) a single network stream using only RGB frames; ii) transfer learning ability and flexibility because of its compatibility with any pretrained state-of-the-art 2D-CNN models for spatial feature extraction; iii) a customisable and simpler architecture compared to existing 3D-CNN and RNN models; and iv) fast learning speed and state-of-the-art performance in HAR.
These benefits allow TransNet to leverage the power and success of transfer learning for HAR, bringing huge advantages in terms of efficiency and effectiveness.

An additional contribution of this paper is that we introduce the strategy of utilising autoencoders to form the TransNet's 2D component, i.e., named {\it TransNet+}, see Figure~\ref{fig:BypassingProcess}.  TransNet+ employs the encoder part of the autoencoder trained on computer vision tasks like human semantic segmentation (HSS) to conduct HAR. Extensive experimental results and the comparison with the state-of-the-art models demonstrate the superior performance of the proposed TransNet/TransNet+ in HAR.

\begin{figure*}
     \centering
\begin{tabular}{c|c}
\includegraphics[scale=0.18]{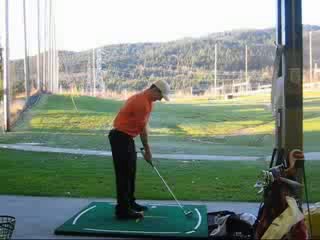}
\includegraphics[scale=0.18]{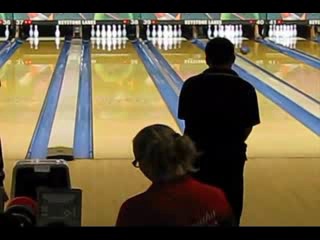}
\includegraphics[scale=0.18]{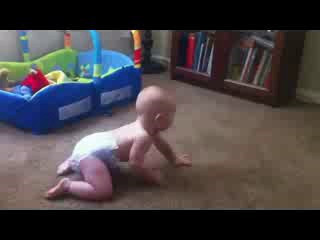}
\includegraphics[scale=0.18]{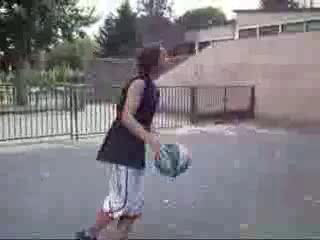} &
\includegraphics[scale=0.18]{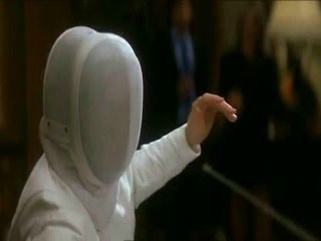}
\includegraphics[scale=0.18]{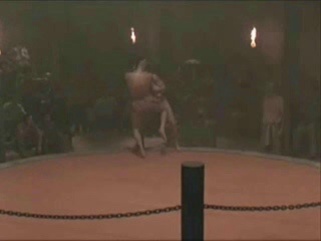}
\includegraphics[scale=0.18]{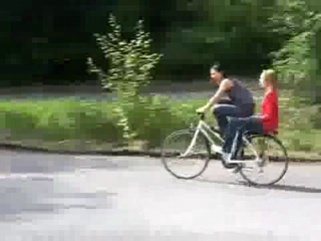}
\includegraphics[scale=0.18]{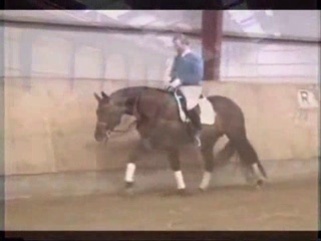}
\end{tabular}
\begin{tabular}{c|c}
\includegraphics[width=1.6cm]{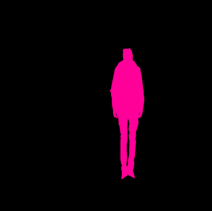}
\includegraphics[width=1.6cm]{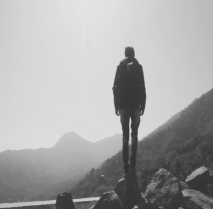}&
\includegraphics[scale=0.19]{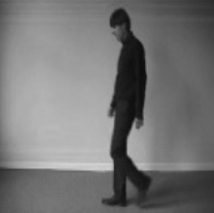}
\includegraphics[scale=0.19]{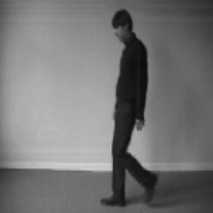}
\includegraphics[scale=0.19]{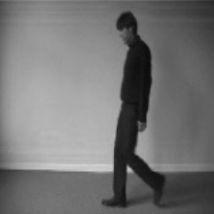}
\includegraphics[scale=0.19]{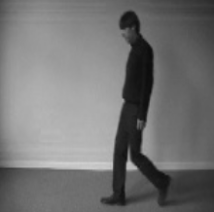}
\includegraphics[scale=0.19]{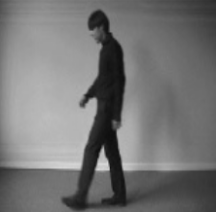}
\includegraphics[scale=0.19]{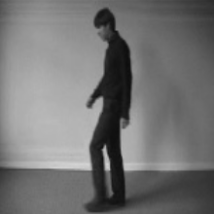}
\includegraphics[scale=0.19]{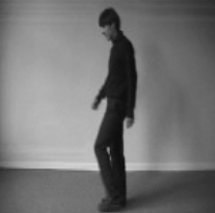}
\includegraphics[scale=0.19]{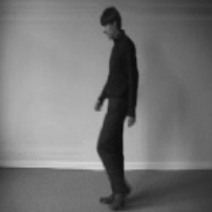}
\end{tabular}
    \caption{Data samples. First row: samples of UCF101 actions (left) and HMDB51 actions (right); second row: samples of the supervisely person dataset (left) and a frame sequence of the action class ``walking" from the KTH dataset (right).}
    \label{fig:data-sample}
\end{figure*}

%-------------------
\section{Related Work} \label{sec:related-work}
%-------------------
%-----------
\subsection{HAR with background subtraction}
%-----------
Most research on HAR focuses on human detection and motion tracking \cite{jaouedi2020new}. Background subtraction has been suggested in a number of methods and proven to be viable for HAR. For example, a background updating model based on a dynamic optimisation threshold method was developed in \cite{zhang2010motion} to detect more complete features of the moving object.
The work in \cite{kim2018hybrid} introduced a basic framework for detecting and recognising moving objects in outdoor CCTV video data using background subtraction and CNNs. Jaouedi et al. \cite{jaouedi2020new} employed a Gaussian mixture model and Kalman filter \cite{liu2007hierarchical} techniques to detect human motion by subtracting the background.

\subsection{HAR with multimodality}
%-----------
Since video comprehension requires motion information, researchers have integrated several input modalities in addition to RGB frames to capture the correlation between frames in an effort to enhance model performance.

{\bf Optical flow.} Optical flow \cite{horn1981bg}, which effectively describes object or scene motion flow, is one of the earliest attempts to capture temporal patterns in videos.  
In comparison to RGB images, optical flow may successfully remove the static background from scenes, resulting in a simpler learning problem than using RGB images as input \cite{diamantas2020optical, wang2018learning}. 
Simonyan et~al. \cite{ryoo2019assemblenet} began the trend of using multiple input modalities, including optical flow, with CNNs.  
However, when compared to the latest deep learning techniques, optical flow has a number of disadvantages, including being computationally complex and highly noise-sensitive \cite{bovik2009essential,li1998neural}, which make its use in real-time applications less feasible.

{\bf Semantic segmentation.}
Semantic segmentation is a technique that may be used to separate either the entire body or particular body parts from the background \cite{minaee2021image}. It is a pixel-wise labelling of a 2D image, offering spatial features describing the shape of the object of interest \cite{ulku2022survey}. Zolfaghari et al. \cite{zolfaghari2017chained} presented a chained multi-stream model that pre-computes and integrates appearance, optical flow, and human body part segmentation to achieve better action recognition and localisation. Benitez et al. \cite{benitez2021improving} offered an alternative to the costly optical flow estimates used in multimodal hand gesture recognition methods. It was built using RGB frames and hand segmentation masks, with better results achieved.

Although semantic segmentation approaches have shown promising outcomes in action recognition, the majority of them are computationally demanding. In fact, real-world action recognition methods involving semantic segmentation of video content are still in their infancy phase \cite{zhang2019comprehensive}.

In sum, most of the aforementioned research focused on creating synthetic images that reflect different input modalities and then analysing them using action recognition models.
Pre-computing multiple input modalities such as optical flow, body part segmentation, and semantic segmentation can be computationally and storage-intensive, making them unsuitable for large-scale training and real-time deployment. Since research in the subject of semantic segmentation may still be in its early stage, one of the objectives of this study is to enhance its potential in HAR.

\subsection{3D-CNNs decomposition}
%-----------
Video can be conceptually simplified by viewing it as a 3D tensor with two spatial dimensions and one time dimension. As a result, 3D-CNNs are adopted to model the spatial and temporal data in video as a processing unit \cite{yao2019review, ji20123d, kalfaoglu2020late}.  Ji et al. \cite{ji20123d}  proposed the pioneer work in the application of 3D-CNNs in action recognition. Although the model's performance is encouraging, the network's depth is insufficient to demonstrate its potential. Tran et al. \cite{tran2015learning} extended the work in \cite{ji20123d} to a 3D network with more depth, called C3D. C3D adopts the modular architecture, which can be viewed as a 3D version of the VGG16 network.

It is worth noting that training a sufficiently deep 3D-CNN from scratch will result in much higher computational cost and memory requirements compared to 2D-CNNs. Furthermore, 3D networks are complex and difficult to optimise \cite{kong2021spatiotemporal}; therefore, a big dataset with diverse video data and activity categories is required to train a 3D-CNN effectively. In addition, it is not straightforward for 3D-CNNs to transfer learning from state-of-the-art pretrained 2D-CNN models since kernel shapes are completely different. Carreira et al. \cite{carreira2017quo} proposed I3D, a 3D-CNN architecture that circumvents the dilemma that 3D-CNNs must be trained from scratch. A strategy was employed to transform the weights of pretrained 2D models, e.g. on ImageNet, to their 3D counterparts. To understand this intuitively, they repeated the weights of the trained 2D kernels along the time dimension of the 3D kernels. Although I3D was successful in overcoming the challenge of spatial transfer-learning, its 3D kernels require enormous quantities of action recognition datasets to capture temporal features. Moreover, the way that I3D stretches 2D-CNN models into 3D-CNNs remains computationally expensive.

P3D \cite{qiu2017learning} and R2+1D \cite{tran2018closer} investigate the concept of decomposing the 3D CNN's kernels into 2D and 1D kernels. They differ in their arrangement of the two factorised operations and their formulation of each residual block. This kind of approach to 3D network decomposition acts at the kernel level. The notion of kernel-level factorisation restricts the ability to switch models (e.g., ResNet50 and VGG16) based on implementation requirements and hinders transfer learning from the current state-of-the-art models.

%-------------------
\section{Proposed TransNet} \label{sec:model}
%-------------------
In this section, we first present our motivations and then introduce the proposed TransNet and its variants.

%-----------
\subsection{Preliminary}
%-----------
%

Video data analysis in deep learning commonly involves two types of approaches: 2D-CNN-RNN \cite{yang2020cnn, fan2016video, abdullah2020facial, rangasamy2020deep} and 3D-CNN \cite{diba2016efficient, hegde2018morph, hou2019efficient}. The CNN-RNN approach comprises a spatial component based on 2D-CNN and a temporal component based on RNN, offering customisation in the 2D-CNN part. However, it often requires longer training time due to the complexity of RNN compared to CNN \cite{prokhorov2002adaptive}. On the other hand, 3D-CNN is faster and simpler to implement but struggles with convergence and generalisation when dealing with limited datasets \cite{wang20203d}. Alternatively, the implementation of 1D-CNN in temporal data analysis holds promise for developing more accurate and efficient models \cite{martin2021three, liu2021data}.

The other main motivation is transfer learning, applying well-designed and well-trained models learnt from one task (i.e., the source task, generally with large data available) to another (i.e., the target task, generally with limited data available) for performance enhancement \cite{zhang2017effect}. The underlying essential assumption is that the source and target tasks are sufficiently similar \cite{zhang2017effect, taylor2009transfer}. In the data scarcity scenario, models may be prone to overfitting, and data augmentation may not be enough to resolve the issue \cite{zhang2021data}. Therefore, transfer learning could play a key role in this regard.

Since HAR requires 3D data analysis, obtaining an optimised model requires training on a large amount of data compared to 2D data \cite{habibie2019wild, hu20213d}. This calls for the use of transfer learning, e.g., pre-training state-of-the-art models first to classify 2D images using large datasets such as ImageNet. However, it is important to study and verify the assumption that the HAR task shares sufficient similarities with the image classification task. Previous research in \cite{geirhos2018imagenet} has shown disparities between CNNs trained on ImageNet and human observers in terms of shape and texture cues, with CNNs exhibiting a strong preference for texture over shape. Additionally, several studies suggest that object shape representations hold greater importance in action recognition tasks \cite{hirota2021grasping, zhang2021video, dhiman2020view, el2018human}.

%-----------
\subsection{Methodology}
\label{Methodology}
%-----------

{\bf TransNet.} We propose to construct a paradigm of utilising the synergy of 2D- and 1D-CNNs, see Figure \ref{fig:2D+1DModel} for the end-to-end {\it TransNet} architecture. TransNet provides flexibility to the 2D-CNN component in terms of model customisability (i.e., using different state-of-the-art 2D-CNN models) and transferability (i.e., involving transfer learning); moreover, it benefits from the 1D-CNN component supporting the development of faster and less complex action recognition models.

TransNet includes the time-distributed layer wrapping the 2D-CNN model. In particular, the 2D component is customisable, and any sought-after 2D-CNN models (e.g., MobileNet, MobileNetV2, VGG16 or VGG19) can be utilised. The time-distributed layer is followed by three 1D convolutional layers for spatio-temporal analysis. In detail, the first one's kernels process the feature map vectors over $(n-1)$ steps, where each kernel has a size of 2, capturing the correlations between a frame and its neighbour, and $n$ is the number of frames in a video clip; the second one's kernels have a size of $(n-1)$, analysing all feature vectors in one step to capture the whole temporal pattern of the frame-sequence; and the third one uses the SoftMax function for the final classification, followed by the flatten layer. More details are given below.

We first define the symbols used for semantic segmentation. Let $\boldsymbol{X}$ represent the input image, and $\boldsymbol{z} = p_\theta(\boldsymbol{X}) \in \mathbb{R}^L$ be the output vector (i.e., latent representation) of the encoder function $p_\theta$ (e.g. MobileNet or VGG16) with parameters $\theta$. The decoder function
is defined analogously. The formed autoencoder can then be trained with the ground truth images.

Let $\mathcal{X}$ be a collection of $n$ frames $\mathcal{X}=\{\boldsymbol{X}^i\}_{i=1}^n$, which is fed into the 2D component (spatial component) of the TransNet architecture in Figure \ref{fig:2D+1DModel}.
The trained encoder $p_{\theta}$ is then used $n$ times to process $\mathcal{X}$ frame by frame and create a set of $n$ spatial feature vectors $\mathcal{Z}=\{\boldsymbol{z}^i\}_{i=1}^n$, where $\boldsymbol{z}^i = p_\theta(\boldsymbol{X}^i)$.
Let $\{\boldsymbol{w}^{j,1},  \boldsymbol{w}^{j,2}\}_{j=1}^K$ be a set of weights, where $\boldsymbol{w}^{j,1}, \boldsymbol{w}^{j,2} \in \mathbb{R}^L$.
The first of the three 1D layers (i.e., the temporal component) processes every two adjacent spatial vectors of $\mathcal{Z}$, i.e., $\{\boldsymbol{z}^i, \boldsymbol{z}^{i+1}\}$, to generate the corresponding spatio-temporal feature vectors $\boldsymbol{h}^i = (h^i_1, \cdots, h^i_K) \in \mathbb{R}^K, i = 1, \ldots, n-1$, where
\begin{equation*} h^i_j = f(\sum_{l=1}^{L}\sum_{k=i}^{i+1}z^k_l w^{j,k-i+1}_l+b^{j}_i), \ \  j=1, \ldots, K,
\end{equation*}
$b^{j}_i$ are the biases and $f$ is the activation function (i.e., Relu $f(x)=\max(0,x)$ is used here).
Let $\{\hat{\boldsymbol{w}}^{j, 1}, \hat{\boldsymbol{w}}^{j, 2}, \cdots, \hat{\boldsymbol{w}}^{j, n-1}\}_{j=1}^C$ be another set of weights, with $\hat{\boldsymbol{w}}^{j,k} \in \mathbb{R}^K, k=1, \ldots, n-1$. The second 1D layer processes the set of spatio-temporal vectors $\{\boldsymbol{h}^{i}\}_{i=1}^{n-1}$ to generate a single spatio-temporal vector $\boldsymbol{v} = (v_1, \cdots, v_C) \in \mathbb{R}^C$, where
\begin{equation*} v_j=f(\sum_{l=1}^{K}\sum_{k=1}^{n-1}h^k_l\hat{w}^{j,k}_l+\hat{b}^j), \ \ j = 1, \ldots, C,
\end{equation*}
and $\hat{b}^j$ are the biases.
Finally, the Softmax layer is used on $\boldsymbol{v}$ to classify action classes.

{\bf TransNet+.} Except for using a sought-after 2D-CNN for TransNet's 2D component, below we present a way of leveraging transfer learning for it. To do so, we construct an autoencoder where TransNet's 2D component serves as its encoder. The autoencoder is then trained on a specific computer vision task such as HSS to extract specific features such as human shape, e.g., see the left of the second row in Figure~\ref{fig:data-sample}. After training, the encoder's parameters become saturated with weights that are capable of describing the features of the required task, such as HAR, see Figure \ref{fig:BypassingProcess}. In this way, the features like object shape that TransNet's 2D component needs to learn can be predetermined by training the autoencoder. We name this way of executing TransNet as {\it TransNet+}.

Note that autoencoders have been used in action recognition challenges e.g. \cite{zolfaghari2017chained}. However, there are a number of disadvantages in their use of autoencoders, including the multiplicity of modelling steps, the need for a large amount of memory, and the lengthy training time due to the high computational cost of training the autoencoder network and action recognition network. 

In contrast, TransNet+ is a huge step further in contributing to the development of an end-to-end HAR model with potential in real-time implementation, since it simplifies the process by just integrating the trained encoder rather than the entire autoencoder in TransNet, with the premise that the trained encoder carries weights capable of describing important features (see Figure \ref{fig:BypassingProcess}).

On the whole, the traditional method of using autoencoders in HAR differs from TransNet+ in that the traditional method uses the entire autoencoder and its output as the next stage's input, whereas TransNet+ just employs the trained encoder of the autoencoder for spatial feature extraction.

{\bf Model complexity.}
The proposed TransNet model is customisable, and thus its size varies depending on the 2D-CNN model being used in the spatial component. In particular, it is quite cost-effective since it uses a time-distributed layer, allowing the 2D-CNN to be used repeatedly without expanding in size. Table \ref{table:TransnetComplexity} gives the number of parameters regarding different choices of the 2D-CNN models.

\begin{table}[t]
\centering
\caption{{TransNet's model complexity. The last column gives the total number of parameters for each setting. The 2D component column reflects the model size of the time-distributed layer, which is invariant against the number of frames in the input video clip. The two values in the 1D component column show the number of kernels in the first and second 1D-CNN layers, respectively. In our experiments, we equipped TransNet with MobileNetV1 and 64 filters.}}
\resizebox{0.46\textwidth}{!}{
\begin{tabular}{c|cccc}
\hline
   2D model  &2D component& 1D component & Total  \\ \hline \hline
   \multirow{2}{*}{MobileNetV1} &6,444,288& 64; 6  & 6,449,416  \\ 
     &9,655,616& 128; 6  &  9,673,992 \\  \hline
 \multirow{2}{*}{MobileNetV2}  &6,277,248& 64; 6 & 6,282,376 \\
 &10,291,392& 128; 6 & 10,309,768\\
\hline
\multirow{2}{*}{VGG16}  &16,322,432& 64; 6 & 16,327,560 \\
 &17,928,128& 128; 6 & 17,946,504 \\ \hline
\multirow{2}{*}{VGG19}  &21,632,128& 64; 6 & 21,637,256\\
 &23,237,824& 128; 6 & 23,256,200\\
   \hline
\end{tabular}}

\label{table:TransnetComplexity}
\end{table}

\begin{table}[t]
\centering
\caption{Results of TransNet with different backbones and different pretraining strategies on the KTH dataset.}
\resizebox{0.41\textwidth}{!}{
\begin{tabular}{c|cc}
\hline
   TransNet backbone                & Pre-training  & Acc.  \\ \hline \hline
 \multirow{3}{*}{MobileNet}  & None  & $94.35 \pm 0.33$    \\ 
    &  ImageNet & \textbf{$\textbf{100.00}  \pm 0.00$}\\ 
   & HSS & \textbf{$\textbf{100.00} \pm 0.00$}   
  \\ \hline
  \multirow{3}{*}{MobileNetV2}  & None  & $88.31 \pm 0.54$    \\
    & ImageNet  & $95.86 \pm 0.41$ \\ 
    & HSS   & $\textbf{96.40} \pm 0.34$
 \\ \hline
\multirow{3}{*}{VGG16} & None  & $90.12 \pm 0.38$    \\ 
   &  ImageNet  & $96.25 \pm 0.43$    \\ 
   &  HSS  & $\textbf{98.01 }\pm 0.48$    \\   \hline
\multirow{3}{*}{VGG19} & None  & $80.06 \pm 0.72$    \\ 
   & ImageNet   & $88.26 \pm 0.51$    \\   
   &  HSS  & $\textbf{94.39} \pm 0.26$    \\   \hline
\end{tabular}}

\label{table:ComparisionResultKTH}
\end{table}

%-------------------
\section{Data}
%-------------------%
In our study, we use two primary groups of benchmark datasets. The first consists of ImageNet and the supervisely person dataset used for transfer learning, while the second consists of the KTH, HMDB51 and UCF101 datasets used for method evaluation (with a split ratio of 80\% and 20\% for training and test, respectively); see below Figure \ref{fig:data-sample} for a brief description and for some samples from these datasets.

%-----------
\subsection{Transfer leaning datasets}
%-----------
\textbf{ImageNet.}
ImageNet \cite{deng2009imagenet} is a famous database consisting of 14,197,122 images with 1000 categories. Since 2010, it has been used in the ImageNet Large Scale Visual Recognition Challenge (ILSVRC).

\textbf{Supervisely person dataset.}
This dataset \cite{superviselydataset2018} is publicly available for human semantic segmentation, containing 5,711 images and 6,884 high-quality annotated human instances. It is available for use in academic research with the purpose of training machines to segment human bodies.

%-----------
\subsection{Human action recognition datasets}
%-----------
\textbf{KTH.}
In 2004, the Royal Institute of Technology introduced KTH, a non-trivial and publicly available dataset for action recognition \cite{schuldt2004recognizing}. It is one of the most standard datasets, including six actions (i.e., walking, jogging, running, boxing, hand-waving, and hand-clapping). Twenty-five different people did each activity, allowing for variation in performance; moreover, the setting was systematically changed for each actor's action, i.e., outdoors, outdoors with scale variation, outdoors with varied clothing, and indoors. KTH includes 2,391 sequences. All sequences were captured using a stationary camera with 25 fps over homogeneous backgrounds.

\textbf{UCF101.}
In 2012, UCF101 \cite{soomro2012ucf101} was introduced as a follow-up to the earlier UCF50 dataset. It is a realistic (not staged by actors) HAR dataset, containing 13,320 YouTube videos representing 101 human actions. It provides a high level of diversity in terms of object appearance, significant variations in camera viewpoint, object scale, illumination conditions, a cluttered background, etc. These video clips are, in total, over 27 hours in duration. All videos have a fixed frame rate of 25 fps at a resolution of $320\times 240$.

\textbf{HMDB51.} HMDB51 \cite{kuehne2011hmdb} was released in 2011 as a realistic HAR dataset. It was primarily gathered from movies, with a small portion coming from public sources such as YouTube and Google videos. It comprises 6,849 videos organised into 51 action categories, each with at least 101 videos.

%-------------------
\section{Experimental Results }
\label{sec:ExperimentalResults}
%-------------------

%-----------
\subsection{Settings}
%-----------
Our model is built using Python 3.6 with the deep learning library Keras, the image processing library OpenCV, matplotlib, and the scikit-learn library. A computer with an Intel Core i7 processor, an NVidia RTX 2070, and 64GB of RAM is used for training and test.

Four CNN models with small sizes (i.e., MobileNet, MobileNetV2, VGG16, and VGG19) are selected as the backbones of TransNet/TransNet+, with parameter numbers of 3,228,864, 2,258,984, 14,714,688, and 20,024,388 (without the classification layers), respectively.

TransNet with each different backbone is implemented in three different ways: i) untrained; ii) being trained on ImageNet; and iii) being trained on HSS using the supervisely person datasetas as encoders. Note that the last way is described in TransNet+. For ease of reference, we drop the `+' sign in the following. The number of 200 epochs is used to train all autoencoders, with a batch size of 24. The models are first trained and evaluated on the KTH dataset. Then the one with the best performance is selected to be evaluated on all the datasets, and compared with the current state-of-the-art HAR models. Each video clip consists of a sequence of 12 frames, and the input modality is RGB with a size of $224\times 224\times 3$.

%%%%%%%%%%%%%%%%%%%%%%%%%%%%%%%%%%%%%%%%%%%%%%%%%%%%%%%%%%%
\begin{table}[t]
\centering
\caption{Results comparison between TransNet and the state-of-the-art methods on the KTH dataset.}
\resizebox{0.41\textwidth}{!}{
\begin{tabular}{c|cc}
\hline
   Method  & Venue & Acc.  \\ \hline \hline
  Grushin et al. \cite{grushin2013robust}& IJCNN '13 &  90.70  \\
  Shu et al.  \cite{shu2014bio}& IJCNN '14 &  92.30  \\
  Geng et al. \cite{geng2016human}& ICCSAE '15 & 92.49  \\
  Veeriah et al. \cite{veeriah2015differential}& ICCV '15 & 93.96  \\
  Zhang et al. \cite{zhang2017human}& ISMEMS '17 & 95.00  \\
  Arunnehru et al. \cite{arunnehru2018human}
  & RoSMa '18& 94.90  \\
  Abdelbaky et al.  \cite{abdelbaky2020human}
  & ITCE '20 & 87.52  \\
Jaouedi et al.  \cite{jaouedi2020new}
  & KSUCI journal '20 & 96.30  \\
Liu et al.  \cite{liu2020construction}
  & JAIHC '20 & 91.93  \\
  HAR-Depth \cite{sahoo2020har} & TETCI '20 & 97.67 \\
  Ramya et al. \cite{ramya2021human} &MTA journal '21& 91.40  \\
  Lee et al. \cite{lee2021video} &CVF '21& 89.40  \\
  Basha et al. \cite{basha2022information} &MTA journal '22& 96.53  \\
  
  Picco et al.  \cite{picco2023high} &NN journal '23& 90.83  \\
  Ye et al.  \cite{ye2023unified} &CVF '23& 90.90  \\
 \hline
TransNet  & - & \textbf{100.00}     
\\   \hline
\end{tabular}}

\label{table:SOTAKTH}
\end{table}

%%%%%%%%%%%%%%%%%%%%%%%%%%%%%%%%%%%%%%%%%%%%
\begin{table}[t]
\centering
\caption{Results comparison between TransNet and the state-of-the-art methods on the UCF101 dataset.}
\resizebox{0.49\textwidth}{!}{
\begin{tabular}{c|cccc}
\hline
   Model             & Pre-training & {Backbone} & Venue  & Acc.  \\ \hline \hline
  DeepVideo \cite{karpathy2014large} & ImageNet & AlexNet &  CVPR '14  & 65.40   \\ 
  Two-stream \cite{simonyan2014two} & ImageNet  &CNN-M& NeurIPS '14 & 88.00 \\ 
  LRCN  \cite{zhu2020comprehensive} 
  & ImageNet  &CNN-M& CVPR '15 & 82.30 

  \\
  C3D \cite{tran2015learning} 
  & ImageNet &VGG16-like& ICCV '15   & 82.30 
  
  \\Fusion % 
  \cite{feichtenhofer2016convolutional} 
  & ImageNet  &VGG16& CVPR '16 & 92.50    
  \\TSN \cite{zhu2020comprehensive} 
  & ImageNet  &BN-Inception& ECCV '16 & 94.00    
     \\ 
  \multirow{2}{*}{I3D \cite{carreira2017quo}} & ImageNet & \multirow{2}{*}{BN-Inception-like} &  \multirow{2}{*}{CVPR '17}  & \multirow{2}{*}{95.60} \\ 
   &  Kinetics400 & &   &  \\
   
  P3D  \cite{zhu2020comprehensive} 
  &  Sports1M   &ResNet50-like& ICCV '17   & 88.60    
\\ ResNet3D \cite{zhu2020comprehensive}    
&  Kinetics400  &ResNeXt101-like&  CVPR '18  & 94.50    

\\HAR-Depth \cite{sahoo2020har} 
  & -  &BiLSTM+AlexNet&TETCI '20&   92.97 
\\ MemDPC \cite{zhu2020comprehensive}        
&  Kinetics400  &R-2D3D& ECCV '20   &    86.10 
\\
TEA \cite{li2020tea} 
  & Imagenet & ResNet50-like &CVPR '20&   96.90
  \\
CVRL  \cite{zhu2020comprehensive}         
&  Kinetics600  &R3D-50& CVPR '21  &   93.40 
\\
\multirow{2}{*}{TDN \cite{wang2021tdn}} 
  & Kinetics400  & \multirow{2}{*}{ResNet} & \multirow{2}{*}{CVF '21} &   \multirow{2}{*}{97.40}
\\ 
  & ImageNet & & &   
\\ 
Multi-Domain \cite{omi2022model} 
  & -  & MDL & IEICE '22&   94.82  \\
 MEST \cite{zhang2022mest} 
  & Imagenet & 2D-CNN & Sensors '22&   96.80 \\

STA-TSN \cite{yang2022sta} 
  & Imagenet & ResNet-LSTM &PloS One '22&   92.80  \\

  FSAN \cite{chen2023two} 
  & Imagenet & 2D-CNN & Sensors '23&   95.68 \\ 
\hline
\textbf{TransNet} &  ImageNet & MobileNet V1& -  & \textbf{98.32}      
\\   \hline
\end{tabular}}

\label{table:SOTAUCF101}
\end{table}

\begin{table}[t]
\centering
\caption{Results comparison between TransNet and the state-of-the-art methods on the HMDB51 dataset.}

\resizebox{0.49\textwidth}{!}{
\begin{tabular}{c|cccc}
\hline
   Model             & Pre-training  & {Backbone}& Venue  & Acc.  \\ \hline \hline
  C3D \cite{tran2015learning} & ImageNet &VGG16-like& ICCV '15   &  56.80  \\ 
 CBT \cite{zhu2020comprehensive} 
 & ImageNet &S3D& arXiv '19   &  44.60  \\ 
 %XDC \cite{alwassel2020self} & Kinetics400 && NeurIPS 2020   &  52.6  \\ 
 DPC \cite{zhu2020comprehensive} 
 & Kinetics400 & R-2D3D& ICCVW '19   &  35.70  \\ 
 SpeedNet %\cite{benaim2020speednet} 
 & Kinetics400 &S3D-G& CVPR '20   &  48.80  \\ 
 MemDPC \cite{zhu2020comprehensive} 
 & Kinetics400 &R-2D3D&  ECCV '20   &  54.50  \\ 
 CoCLR \cite{zhu2020comprehensive} 
 & Kinetics400 &S3D& NeurIPS '20   &  54.60  \\ 
 HAR-Depth \cite{sahoo2020har} 
  & -  &BiLSTM+AlexNet&TETCI '20&   69.74 
   \\
   STA-TSN \cite{kwon2020motionsqueeze} 
  & Imagenet & ResNet50-like &ECCV '20&   77.40 \\
 TEA \cite{li2020tea} 
  & Imagenet & ResNet50-like &CVPR '20&   73.30
  \\
  \multirow{2}{*}{TDN \cite{wang2021tdn} }
  & Kinetics400 & \multirow{2}{*}{ResNet} & \multirow{2}{*}{CVF '21} &  \multirow{2}{*}{76.30} 
  \\
  & ImageNet & & &   
  \\
  Multi-Domain \cite{omi2022model} 
  & -  &MDL&IEICE '22&   71.57  \\
MEST \cite{zhang2022mest} 
  & Imagenet & 2D-CNN & Sensors '22&   73.40 \\
   STA-TSN \cite{yang2022sta} 
  & Imagenet & ResNet-LSTM &PloS One '22&   68.60 
  \\  
FSAN \cite{chen2023two} 
  & Imagenet & 2D-CNN & Sensors '23&   72.60 \\ 
 \hline
TransNet &  ImageNet &MobileNet V1& -  & \textbf{97.93}     
\\  \hline
\end{tabular}}

\label{table:SOTAHMDB51}
\end{table}

\subsection{Results and discussion}
%-----------
In a nutshell, we conduct experiments below with three main objectives: i) determining whether or not the proposed TransNet architecture can offer a reliable mechanism by leveraging transfer learning; ii) evaluating if the HSS-trained TransNet provides superior spatio-temporal characteristics for HAR than the ImageNet-trained TransNet;  and iii) validating if the performance of the TransNet architecture can achieve state-of-the-art performance in comparison to current state-of-the-art methods in HAR.

Initially, we subject TransNet to an evaluation using the KTH dataset, which serves as a suitable choice due to its primary emphasis on human action detection while excluding the presence of additional objects in the background, in contrast to the UCF101 and HMDB51 datasets. The purpose of this evaluation is to validate the viability of employing HSS as a means of pretraining to improve the performance of the model in similar tasks.

The results presented in Table \ref{table:ComparisionResultKTH} demonstrate the superior performance of the TransNet model which was trained using HSS in comparison to its untrained and ImageNet-trained counterparts. Specifically, the untrained MobileNet, MobileNetV2, VGG16, and VGG19-based TransNet models achieved an average accuracy of 88.21\%, and the ImageNet-trained models achieved an average accuracy of 95.09\%. In contrast, the HSS-trained TransNet models achieved an average accuracy of 97.20\%, indicating a significant improvement of {$\sim8.99\%$} and {$\sim2.11\%$} over the untrained and ImageNet-trained models, respectively. These findings underscore the effectiveness of the pretraining strategy employing autoencoders in enhancing the performance of the TransNet model. Additionally, the findings show the significance of incorporating transfer learning as a means of enhancing performance, thereby bestowing a substantial advantage to the 2D-1D-CNN architecture and enabling us to leverage transfer learning within the 2D-CNN component.

Tables \ref{table:SOTAKTH}, \ref{table:SOTAUCF101} and \ref{table:SOTAHMDB51} present the quantitative comparisons between TransNet and the current state-of-the-art methods being applied to the HAR datasets, i.e., KTH, UCF101 and HMDB51. In these comparisons, a MobileNet-based TransNet  pretrained on ImageNet  is used. The findings demonstrate the exceptional performance achieved by the proposed TransNet, surpassing the existing state-of-the-art results by a considerable margin. Additionally, these findings solidify the 2D-1D-CNN architecture as a highly effective approach for HAR.
%%%%%%%%%%%%%%%%%%%%%%%%%%%%%%%%%%%%%%%%%%%%%%%%%%%%%%%%%%%%

%-------------------
\section{Conclusion}
%-------------------
In this paper, we proposed TransNet, a versatile and highly efficient end-to-end deep learning architecture designed specifically for HAR. TransNet employs a concise architecture, combining 2D- and 1D-CNNs to effectively capture spatial features and temporal patterns in video data, respectively. It demonstrates excellent compatibility and flexibility by seamlessly integrating with various pretrained state-of-the-art 2D-CNN models, thereby facilitating straightforward transfer learning for HAR tasks and significantly improving performance in terms of both efficiency and effectiveness. Through extensive experiments conducted on diverse HAR benchmark datasets utilising different backbones and pre-training strategies, the proposed TransNet consistently outperforms state-of-the-art HAR models. These experiments validated the superior performance of TransNet in HAR, further establishing its efficacy and robustness in the field. The application of transfer learning using autoencoder in the context of HSS has emerged as a promising approach, offering a potential avenue for directing the model to learn task-specific features in accordance with the requirements of the given task. Future work may focus on detailed ablation study of the architecture of TransNet and its combination with other advanced architectures like transformers for HAR.
{\small
\bibliographystyle{IEEEtran}
\bibliography{IEEEabrv}
}

\end{document}